\def\year{2018}\relax
\documentclass[letterpaper]{article} 
\usepackage{aaai18}  
\usepackage{times}  
\usepackage{helvet}  
\usepackage{courier}  
\usepackage{url}  
\usepackage{graphicx}  
\usepackage{amsmath}
\usepackage{mathrsfs}
\usepackage{algorithm}
\usepackage{algorithmic}
\begin{document}
%
\title{Metric Learning-based Generative Adversarial Network}
\author{Zi-Yi Dou\\
National Key Laboratory for Novel Software Technology, Nanjing University\\
Nanjing, 210023, China\\
141242042@smail.nju.edu.cn \\
}
\maketitle
\begin{abstract}
Generative Adversarial Networks (GANs), as a framework for estimating generative models via an adversarial process, have attracted huge attention and have proven to be powerful in a variety of tasks. However, training GANs is well known for being delicate and unstable, partially caused by its sigmoid cross entropy loss function for the discriminator. To overcome such a problem, many researchers directed their attention on various ways to measure how close the model distribution and real distribution are and have applied different metrics as their objective functions. In this paper, we propose a novel framework to train GANs based on distance metric learning and we call it Metric Learning-based Generative Adversarial Network (MLGAN). The discriminator of MLGANs can dynamically learn an appropriate metric, rather than a static one, to measure the distance between generated samples and real samples. Afterwards, MLGANs update the generator under the newly learned metric. We evaluate our approach on several representative datasets and the experimental results demonstrate that MLGANs can achieve superior performance compared with several existing state-of-the-art approaches. We also empirically show that MLGANs could increase the stability of training GANs.
\end{abstract}
\noindent

\section{Introduction}

Generative Adversarial Networks (GANs) are a powerful class of deep generative models  \cite{gan}. The basic strategy of GANs is to train a generative model and a discriminative model simultaneously via an adversarial process: the goal of the generator is to capture the data distribution whereas the discriminator tries to distinguish between the generated samples and real samples.
 GANs have attracted great attention due to their impressive performance on a variety of tasks, such as 
image generation \cite{nguyen2016plug}
,
image super-resolution
\cite{ledig2016photo}
and
semi-supervised learning
\cite{salimans2016improved}
. 
Recent studies have also shown their great capabilities in feature extraction
\cite{radford2015unsupervised}
,
and classification tasks
\cite{salimans2016improved}
.

Despite all the great progress, we should be aware that there are also several limitations for GANs. For example, it is well known that GANs are often hard to train and much of the recent work has been devoted to finding ways
of stabilizing training 
\cite{gulrajani2017improved}
.
In addition, Arjovsky {\it et al.} point out that: 1) in theory, one would expect we would first train the discriminator to optimality and then update the generator. In practice, however, as the discriminator gets better, the updates to the generator get consistently worse. 2) a popular fixation using a generator gradient updating with $E_{z \sim P_Z} [-\log D(G(z))]$ is unstable because of the singularity at the denominator when the discriminator is accurate
\cite{arjovsky2017towards}
.

For GANs, traditional approaches to generative modeling relied on maximizing likelihood, or Jensen-Shannon (JS) divergence between unknown data distribution and generator's distribution. As has been pointed out by several papers
\cite{wgan} \cite{metz2016unrolled} \cite{qi2017loss}
, minimizing such objective function could make GANs suffer from vanishing gradients and thus partially leads to the instability of GANs learning.

To resolve the aforementioned issue, some researchers have directed their attention on various ways to measure how close the model distribution and real distribution are and try to use different metrics as their objective functions to improve the training of GANs. For instance, some papers have found out that applying Wasserstein-1 metric \cite{wgan} or energy distance \cite{bellemare2017cramer} into GANs could enhance the performance of GANs as well as increase the stability of training GANs. Inspired by their work, in this paper we propose an alternative approach, the discriminator of which can dynamically learn an appropriate metric, rather than a static one, to measure the distance between generated images and real images. Afterwards, we update the generator under the newly learned metric. To do so, we borrow the idea from distance metric learning, a field which is mainly concerned with learning a distance function tuned to a particular task. Basically, now the responsibility of the ``discriminator" is to learn an appropriate metric, rather than distinguish between real and fake samples; also, the generator aims at minimizing the distance that has been learned by the ``discriminator". We hope that in this new adversarial framework, we could get better performance as well as more stability.

The contribution of this paper could be listed as follows:

\begin{itemize}
\item We define a novel form of GAN named Metric Learning-based GAN (MLGAN) whose discriminator can dynamically learn a suitable metric and provide a reasonable objective function
for the generator.
\item We empirically show that MLGANs have the capacity to stabilize the training of GANs and meanwhile achieve superior performance compared with several other existing state-of-the-art GAN models.
\end{itemize}

\section{Related Work}
\subsection{Generative Adversarial Network}

A generative algorithm models how the data was generated in order to categorize a signal. Generally, to train a generative model we first need to collect a large amount of data, and then train a model to generate data like it. Before GANs, there are several other generative models.
For example, Restricted Boltzmann Machines (RBMs) 
\cite {smolensky1986information}
\cite{hinton2006fast}
have been used effectively in modeling distributions over binary-valued data and are the basis of many other deep generative models, such as Deep Belief Networks (DBNs) \cite{hinton2009deep} or Deep Boltzmann Machines (DBMs) \cite{salakhutdinov2009deep}. 
DBNs can be formed by stacking RBMs and optionally fine-tuning the resulting deep network with gradient descent and back-propagation
and DBMs are undirected graphical models whose component modules are also RBMs
.
Variational Autoencoder (VAE) is another important generative model, which inherits autoencoder architecture, but make strong assumptions concerning the distribution of latent variables
\cite{kingma2013auto}
. 

In 2014, Goodfellow {\it et al.} proposed a new framework for estimating generative models via an adversarial process \cite{gan} and have attracted huge attention due to their promising results in many fields, like text to image synthesis 
\cite{reed2016generative}
and image to image translation
\cite{isola2016image}
.
Unlike aforementioned deep generative models, GANs do not require any approximation method and offer much more flexibility in the definition of the objective function. Also, the goal of GANs, which is generate data that is indistinguishable from data by the discriminator, is highly aligned with the goal of producing realistic data. However, we should also be aware that training GANs is well known for being delicate and unstable as we have mentioned before
\cite{wgan}
.

To alleviate the problem, Arjovsky {\it et al.} propose Wasserstein GAN (WGAN), which use Earth-Mover (also called Wasserstein-1) distance as their objective function
\cite{wgan}
. However, to enforce the Lipschitz constraint on the critic, Arjovsky {\it et al.} use a method called weight clipping to clamp the weights of neural networks to a fixed box
. 
Gulrajani {\it et al.} argue that weight clipping in WGAN could lead to optimization difficulties
\cite{gulrajani2017improved}
. To resolve the issue, they add gradient penalty in their objective function to provide an alternative way to enforce the Lipschitz constraint. 
In addition to WGAN, Mao {\it et al.} propose Least Squares GANs (LSGANs) \cite{mao2016least} which adopt the least squares loss function for the discriminator and Zhao {\it et al.} propose Energy-Based GANs (EBGANs) \cite{zhao2016energy} which view the discriminator as an energy function that attributes low energies to the regions near the data manifold and higher energies to other regions. 

Different from the above variants of GANs, which could be viewed as minimizing a static divergence between the real distribution and the generator's distribution, our method could dynamically learn a suitable metric to measure the difference between them and thus provide the generator with a more reasonable objective function.

\subsection{Distance Metric Learning}

The basic idea of distance metric learning is to find a
distance metric such that the distance between data points in the same
class is smaller than that from different classes
\cite{ye2016instance}
.
To achieve this goal, different methods use various criteria.
For example, 
Xing {\it et al.} 
pose metric learning as a constrained convex optimization problem \cite{xing2003distance} and Goldberger {\it et al.}
propose Neighborhood Components Analysis (NCA) whose main idea is to optimize a softmax version of the leave-one-out K-Nearest-Neighbor (KNN) score
\cite{goldberger2005neighbourhood}
.
There are many other methods and more information about metric learning could be found in 
\cite{bellet2013survey}
\cite{kulis2013metric}
.

With the success of deep learning, deep metric learning has gained much popularity in recent years.
Compared to previous distance metric learning approaches, deep metric learning learns a nonlinear embedding of the data by using deep neural networks.
The approach of Chopra {\it et al.} 
\cite{chopra2005learning}
considers utilizing convolutional neural networks to learn a similarity metric using contrastive loss and   
Hoffer {\it et al.} propose the triplet network model which aims to learn useful representations by distance comparisons
\cite{hoffer2014deep}
.

Recently, Zieba {\it et al.} put forward a method to train a triplet network by putting it as the discriminator in GANs
\cite{zieba2017training}
.
They make use of the good capability of representation learning of the discriminator to increase the predictive quality of the model.
Contrary to their work, whose goal is to enhance the performance of models in deep metric learning, in this paper our method aims at improving the performance and stability of GANs. 

\section{Proposed Method}
In this section, we first illustrate the symbols and definitions that will be used in the following part, and then briefly introduce regular GANs and MMC, a well-known method in distance metric learning. Then, we describe the basic framework of our proposed method (MLGAN). At last, we present two improvements to MLGANs that could make MLGANs achieve better performance.

\begin{figure*}[h!]
\centering
\includegraphics[width=\textwidth]{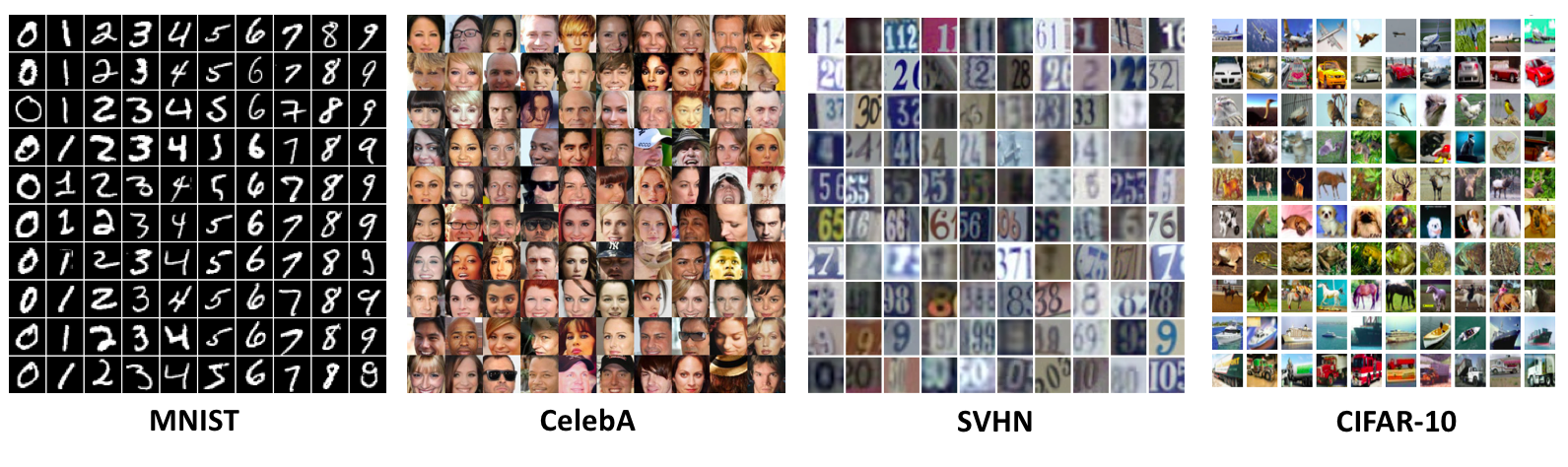}
\caption{\label{real} Real images of four datasets}  
\end{figure*}
\subsection{Symbols and Definitions}
We denote the generator as $G(\mathbf{z};\theta_g)$ and discriminator as $D(\mathbf{x};\theta_d)$ and in this paper they are both deep convolutional neural networks.
To learn the generator's distribution $p_g$ over data $\mathbf{x}$, we define a prior on input noise variables $p_z(\mathbf{z})$, then represent a mapping to data space
as $G(\mathbf{z};\theta_g)$. The discriminator $D(\mathbf{x};\theta_d)$, on the other hand, outputs a vector for each data $\mathbf{x}$. It should be noted that compared to 
regular GANs, in this paper $D$ outputs a vector rather than a single scalar and $D(\mathbf{x};\theta_d)$ represents an embedding of $\mathbf{x}$ instead of the probability that $\mathbf{x}$ came from read distribution $p_{data}$ rather than $p_g$. In this way, it may be a little inappropriate to call it a ``discriminator", but we decide to still use the word for consistency.

During the training of MLGANs, we would sample minibatch of $m$ real examples $\{x^{(i)}\}^m_{i=1}$ $\sim$ $p_{data}(x) $ for each epoch, which would be denoted as $B_{data}$, and also minibatch of $m$ noise samples $\{z^{(i)}\}^m_{i=1}$ $\sim$ $p_g(z) $ for each epoch, which would be denoted as $B_{g}$.

\subsection{Regular GANs}

The GANs training strategy is to define a game between two competing networks. The generator network $G$ maps a source of noise to the input space. The discriminator network $D$ receive either a generated sample or a true data sample and must distinguish between the two. The generator is trained to fool the discriminator.

Formally, the game between the generator $G$ and the discriminator $D$ is the minimax objective:
\begin{equation}
\label{e1}
\min\limits_{G} \max\limits_{D} \mathop{E}\limits_{x\sim p_{data}} \log (D(x)) + \mathop{E}\limits_{z\sim p(z)} \log (1- D(G(z)))
\end{equation}

\subsection{MMC}

As we have stated above, the main idea of MLGAN is to dynamically learn an appropriate metric, rather than a static one, to measure the distance between generated images and real images. After obtaining the metric, we update the generator $G$ under the learned metric.
To do so, we borrow the idea from distance metric learning.

 In the metric learning literature, the term ``Mahalanobis distance" is often used to denote any distance function of the form
\begin{equation}
\label{eq1}
d_A(x,y) = (x-y)^T A(x-y),
\end{equation}
where A is some positive semi-definite matrix. Since A is positive semi-definite, we factorize it as $A=G^TG$ and simple algebraic manipulations would show that 
\begin {equation}
\label{eq2}
 d_A(x,y) = || Gx-Gy||_2^2.
 \end{equation}
Thus, this generalized notion of a Mahalanobis distance exactly captures the idea of learning a global linear transformation.

As has been discussed before, there are several existing works on metric learning and one of the most famous methods was proposed by Xing {\it et al.}
\cite{xing2003distance}
, sometimes referred to as MMC. The main idea of MMC is to minimize the sum of distances that should be similar while maximizing the sum of distances that should be dissimilar.
In MMC's setting, they have some points $\{x_i\}_{i=1}^m \in R^n$ and are given information that
certain pairs of them are ``similar":
$$
\emph{S}:(x_i, x_j) \in \emph{S} , \text{if } x_i \text{ and } x_j \text{ are similar}
$$

A simple way of defining a criterion for the desired metric is to demand that pairs of points $(x_i, x_j)$ in $S$ have small squared distance between them. In order to ensure that $A$ does not collapse the dataset into a single point, they also add a constraint that the pairs of points $(x_i, x_j)$ in $\emph{DS}$, which means they are known to be 
dissimilar, should be separated.
This gives the following optimization problem for MMC:
\begin{equation}
\begin{aligned}
\mathop{\min}\limits_{A} & \sum\limits_{(x_i, x_j) \in \emph{S}} d_A(x_i, x_j) \\
s.t. & \sum\limits_{(x_i, x_j) \in \emph{DS}} \sqrt{d_A(x_i, x_j)} \geq 1 \\
& A \succeq 0
\end{aligned}
\end{equation}

The authors utilize $\sqrt{d_A(x_i, x_j)} $ instead of the usual squared Mahalanobis distance.
The authors also discuss that in the case that we want to learn a diagonal $A=diag(A_{11}, A_{22}, ..., A_{nn})$, we can derive an efficient algorithm using the Newton-Raphson method.
Define
\begin{equation}
\begin{aligned}
g(A)&=g(A_{11}, \cdots, A_{nn})  \\
 &= \sum\limits_{(x_i, x_j) \in \emph{S}} d_A(x_i, x_j) 
 - \log(\sum\limits_{(x_i, x_j) \in \emph{DS}} \sqrt{d_A(x_i, x_j)})
\end{aligned}
\end{equation}

It is straightforward to show that minimizing $g$ (subject to $A \succeq 0$ is equivalent, up to a multiplication of $A$ by
a positive constant, to solving the original problem.
\begin{figure*}[h!]
\centering
\includegraphics[width=\textwidth]{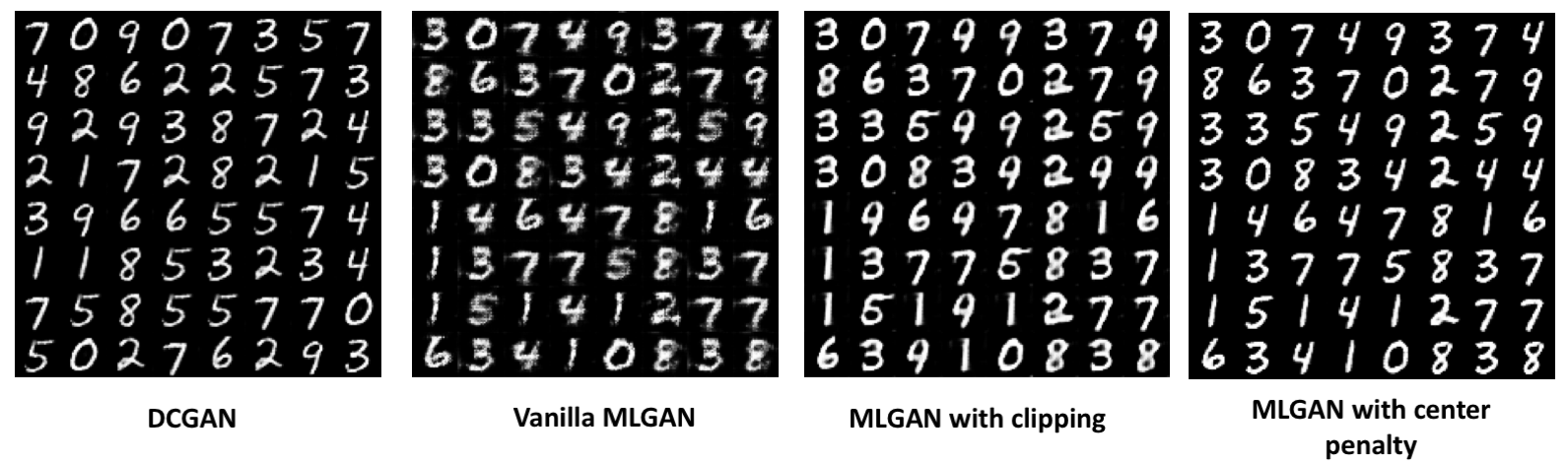}
\caption{\label{f1} Experimental results on MNIST}  
\end{figure*}
\begin{figure*}[h!]
\centering
\includegraphics[width=\textwidth]{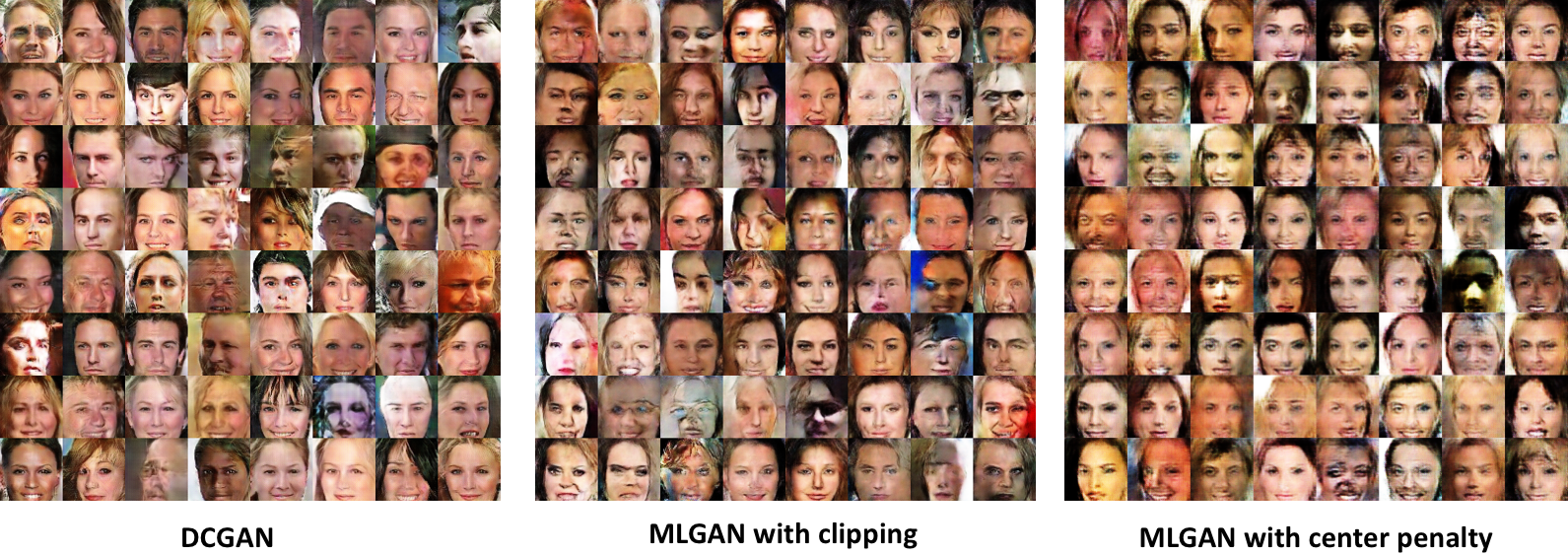}
\caption{\label{f2} Experimental results on CelebA}  
\end{figure*}
\subsection{Basic Framework of MLGAN}
As we have stated above, MMC could be viewed as 
finding a rescaling of a data that replaces each point $x$ with
$Gx$ and then applying standard Euclidean metric to the rescaled data.
This idea could be easily extended to nonlinear metric learning.

To learn a nonlinear metric, we could utilize artificial neural networks. To illustrate, the equation \ref{eq2} could be changed to 
\begin {equation}
\label{eq3}
 d(x,y) = || \phi(x)-\phi(y)||_2^2,
 \end{equation}
and in order to parameterize the mapping $\phi$, we could consider learning a deep neural network.

In this paper, inspired by MMC and their objective function for the diagonal case, we propose a novel objective function for the discriminator of MLGAN.

First, we define 
\begin{equation}
\begin{aligned}
L_{intra} &=   \sum\limits_{(x_i, x_j) \in S_{data}} ||D(x_i) - D(x_j)||_2 ^2 \\ &+  \sum\limits_{(G(z_i), G(z_j)) \in S_g}   ||D(G(z_i)) - D(G(z_j))||_2 ^2
\end{aligned}
\end{equation}
, where 
$
S_{data}=\{  (x_i, x_j) | x_i, x_j \in B_{data}  \}
$
and 
$
S_{g}=\{  (G(z_i), G(z_j)) | z_i, z_j \in B_g \}.
$

We can see that $L_{intra}$ represents the sum of distance within each class.

We also define 
\begin{equation}
\begin{aligned}
L_{inter} &= \sum\limits_{(x_i, G(z_i)) \in DS}   {||D(x_i) - D(G(z_i))||_1}
\end{aligned}
\end{equation}
, where 
$
DS=\{  (x_i, G(z_i)) | x_i \in B_{data}, z_i \in B_g   \}.
$

Again, we can see $L_{inter}$ represents the sum of distance between classes.

Finally, we can present the objective function for the discriminator
\begin{equation}
\label{d_ob}
\begin{aligned}
\centering
L_d = \min\limits_{\theta_d} L_{intra} - \lambda * L_{inter}.
 \end{aligned}
\end{equation}
Here $\lambda$ is a hyper-parameter that balances the two terms.

And on the other hand, the objective function for generator is simply:
\begin{equation}
\label{g_ob}
\begin{aligned}
\centering
L_g =&\min\limits_{\theta_g} L_{inter}.
 \end{aligned}
\end{equation}

Basically, for the discriminator minimizing the first term $L_{intra}$ in the objective function imply that each real data $x$ in $S_{data}$ should be similar to each other and so does each fake data generated by $G$. Also, minimizing the second term $-L_{inter}$ of the objective function means that the real data should be dissimilar to the generated data. Based on this idea, the discriminator $D$, which is a deep neural network, embeds the original data $x$ into $D(x)$ so that in the new embedding 
space the standard distance between them satisfy the aforementioned condition. We have also tried several other variants of objective function for both discriminator and generator, but we find those variants lead to worse performance.  

In regular GANs, the goal of generator $G$ is to fool the discriminator $D$ so that $D$ cannot distinguish between real or generated samples. In this work, however, the generator $G$ is trained to generate samples that is close to the real data under the newly learned metric. Intuitively, in this way the discriminator $D$ could inform the generator $G$ where it should pay attention to correct itself, and then the generator $G$ would try to fix its mistake based on the information told by the discriminator $D$.

To summarize, the main differences between regular GANs and MLGANs are as follows:
\begin{itemize}
\item The objective function for regular GANs is Equation \ref{e1} whereas for MLGANs the objective functions are Equation \ref{d_ob} and Equation \ref{g_ob}.
\item The discriminator $D$ of MLGAN does not have a {\it softmax} layer.
\item The discriminator $D$ of MLGAN could output a real vector for each data $x$ rather than a single scalar.
\item When training the generator, MLGAN still needs to use the minibatch of real data.
\end{itemize}

The whole procedure of proposed algorithm is illustrated in Algorithm 1. It should be noted that since we would make some improvements to MLGANs, which would be illustrated in later part, the revised procedure might be slightly different from Algorithm 1, but the basic framework would be the same.
\begin{algorithm}
\caption{Vanilla MLGAN}
\textbf{Input:} The number of critic iterations per generator iteration $n_{critic}$, the batch size $m$, Adam hyper-parameters $\alpha, \beta_1, \beta_2$ \\
\begin{algorithmic}[1]
\WHILE{ $\theta$ has not converged}
	\FOR{$t=0, \cdots, n_{critic}$}
		\STATE{Sample minibatch of $m$ examples $\{x^{(i)}\}^m_{i=1}$ $\sim$ $p_{data}(x) $ }
		\STATE{Sample minibatch of $m$ noise samples $\{z^{(i)}\}^m_{i=1}$ $\sim$ $p_g(z) $ }
		\STATE{ $grad_{\theta_d} = \mathop{\nabla}_{\theta_d} L_d$ }
		\STATE{$\theta_d$ = Adam($ grad_{\theta_d}, \theta_d, \alpha, \beta_1, \beta_2$)   }
	\ENDFOR
	\STATE{Sample minibatch of $m$ examples $\{x^{(i)}\}^m_{i=1}$ $\sim$ $p_{data}(x) $}
	\STATE{Sample minibatch of $m$ noise samples $\{z^{(i)}\}^m_{i=1}$ $\sim$ $p_g(z)$}
	\STATE{ $grad_{\theta_g} = \mathop{\nabla}_{\theta_g} L_g$ }
	\STATE{$\theta_g$ = Adam($ grad_{\theta_g}, \theta_g, \alpha, \beta_1, \beta_2$)   }

\ENDWHILE
\end{algorithmic}
\end{algorithm}

\subsection{Two Improvements to MLGANs}
Since the discriminator $D$ does not have any {\it softmax} layer, we may give MLGANs too much ``freedom" since there is no constraint on the output value of $D$. In practice this feature does lead MLGANs to generate unsatisfactory results, which would be shown in the next section. In order to fix the issue, we propose two possible constraints that could improve the performance of MLGANs and they both achieve good results. 

The first improvement we use is ``weight clipping", which has been used in WGAN. To be specific, we clamp the weights of discriminator $D$ to a fixed box so that it could only output value in a certain range.

The second improvement we use is to add two terms called ``center penalty", where we give the real and fake data two center vectors $\mu_{data}$ and $\mu_{g}$. The discriminator $D$ would be punished if it learns an inappropriate embedding for an image away from its center vector. 
The loss function of  ``center penalty" is:
\begin{equation}
L_{center} = \sum\limits_{x_i \in B_{data}} ||x_i - \mu_{data}||_2 ^2  +\sum\limits_{z_i \in B_{g}} ||G(z_i) - \mu_{g}||_2 ^2
\end{equation}
Therefore, now the objective function for $D$ turns into:

\begin{equation}
\label{center}
\begin{aligned}
\centering
L_d = &\min\limits_{\theta_d}  L_{intra} - \lambda L_{inter} + \beta L_{center}.
 \end{aligned}
\end{equation}

And the objective function for the generator will remain the same.

\begin{figure*}[h!]
\centering
\includegraphics[width=\textwidth]{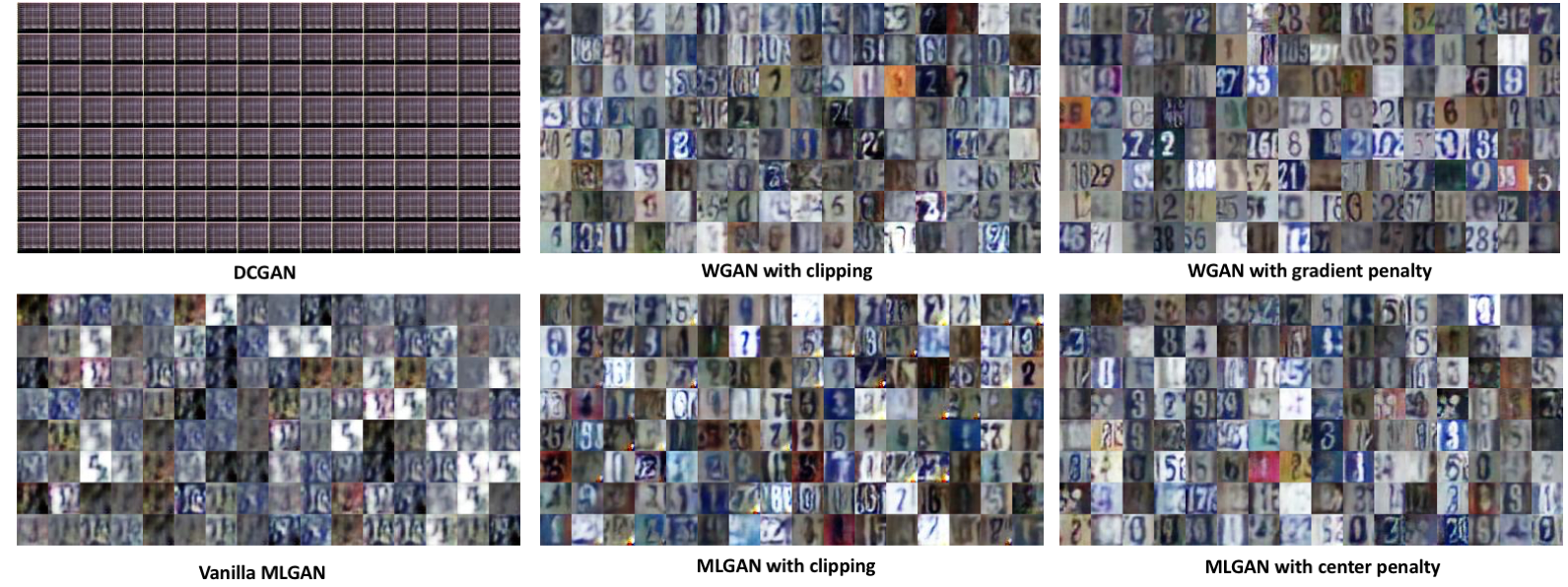}
\caption{\label{f3} Experimental results on SVHN}  
\end{figure*}

\section{Experiment}
\subsection{Datasets and Implementation Details}
We trained MLGANs on four benchmark datasets MNIST
\cite{lecun1998mnist}
, CelebFaces Attributes Dataset (CelebA) 
\cite{liu2015deep}, Street View House Numbers (SVHN) 
\cite{netzer2011reading}
and CIFAR-10
\cite{krizhevsky2009learning}
. The real images of these dataset is shown in Figure \ref{real}

For our experiments, we set  $\lambda$ in Equation \ref{d_ob} to $\frac{1}{2}$ for vanilla MLGAN and MLGAN with clipping, and $1$  for MLGAN with center penalty. The dimension of the output vector of discriminator $d_{dim}$ could be set to $64$ for vanilla MLGAN and MLGAN with clipping, and $5$ for MLGAN with center penalty. $\mu_{data}$ is set to $(\frac{1}{d_{dim}}, \frac{1}{d_{dim}}, \cdots, \frac{1}{d_{dim}})$ and $\mu_{g}$ is set to $(0, 0, \cdots, 0)$. For MLGANs with weight clipping, the clipping threshold is set to $[-0.01, 0.01]$. For MLGANs with center penalty, $\beta$ in Equation \ref{center} is set to $10$ or $20$.

It is difficult to compare performance of different models since GANs lack an objective function. 
Salimans {\it et al.} propose an automatic method to evaluate samples which is now considered as a sound way to assess image quality
\cite{salimans2016improved}
.
Basically, they apply the {\it Inception model} 
\cite{szegedy2016rethinking}
to every generated image to get the conditional label distribution $p(y|x)$. Since images that contain meaningful objects should have a conditional label distribution $p(y|x)$ with low entropy and the marginal $\int p(y|x=G(z)) dz$ with high entropy, their proposed metric  is : $\exp(E_x KL(p(y|x) || p(y)))$. This metric is named {\it Inception score}. In this paper, we utilize {\it Inception score} to compare MLGANs with other models in SVHN and CIFAR-10.
\subsection{MLGAN on MNIST}
The MNIST database \cite{lecun1998mnist} of handwritten digits has a training set set $60,000$ examples and a test set of $10,000$ examples. For MNIST, we use the baseline DCGAN architecture
\cite{radford2015unsupervised}
and our code is based on TensorFlow\cite{abadi2016tensorflow} implementation of DCGAN, whose code is public available
, \footnote{https://github.com/carpedm20/DCGAN-tensorflow}.

The experimental results are shown in Figure \ref{f1}. As we could see from the figure, although the vanilla MLGAN demonstrate comparatively poor performance. After performing the improvement, MLGAN could generate more realistic images.
\begin{figure*}[h!]
\centering
\includegraphics[width=\textwidth]{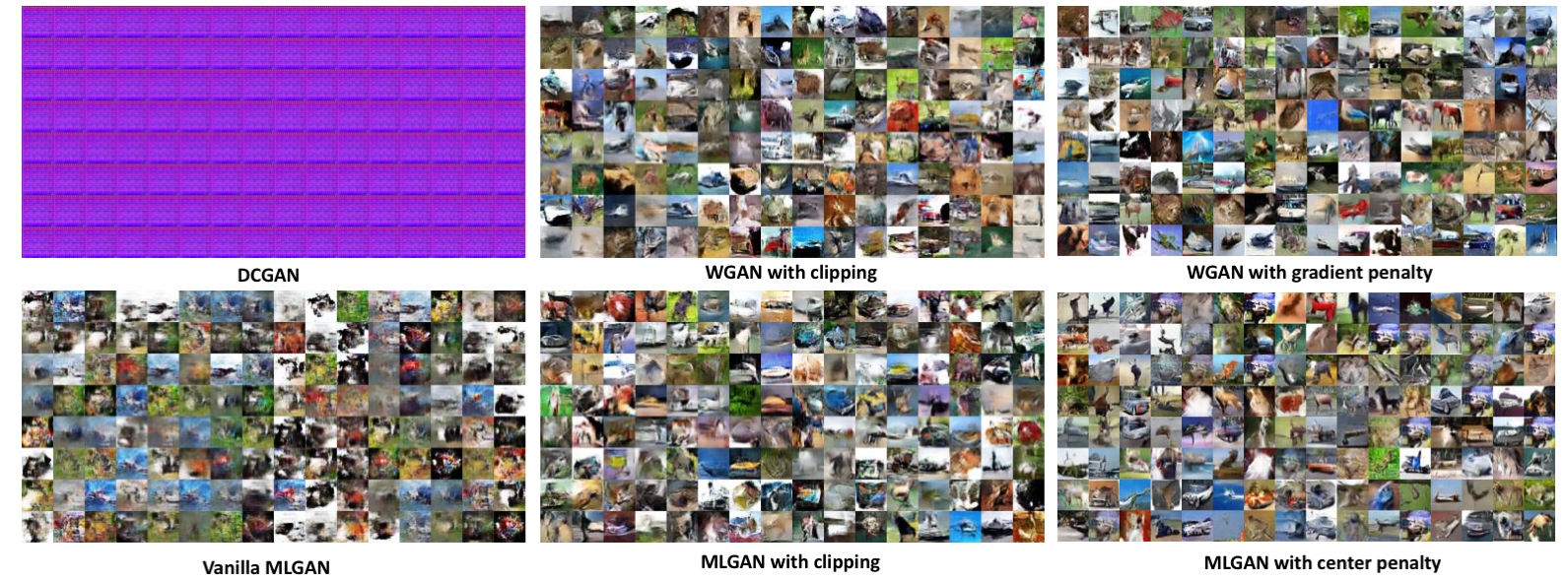}
\caption{\label{f4} Experimental results on CIFAR-10}  
\end{figure*}
\subsection{MLGAN on CelebA}
CelebA \cite{liu2015deep} is a large-scale face
attributes dataset with more than $200$K celebrity images. Figure \ref{f2} shows the comparison of CelebA samples generated by DCGAN and MLGANs. The results demonstrate that MLGANs could achieve competitive performance on CelebA.

\subsection{MLGAN on SVHN}
SVHN \cite{netzer2011reading} is a real-world image dataset for developing machine learning and object recognition algorithms with minimal requirement on data preprocessing and formatting. It can be seen as similar in flavor to MNIST but incorporates and order of magnitude more labeled data and it is obtained from house numbers in Google Street View images.

We use the training set of SVHN, which consists of $73,257$ digits, to train our algorithm and use {\it Inception score} to compare the results with different GAN models.
We use the same model architecture with WGAN with gradient penalty, whose code is public available
\footnote{https://github.com/igul222/improved\_wgan\_training}
.

\begin{table}[h]

\caption{\label{t1}Experimental Results on SVHN}
\begin{tabular*}{8cm}{cc}
\hline
Model (same architecture) & Inception Score\\
\hline
DCGAN & $1.032\pm 0.01$\\
\textbf{MLGAN-vanilla} & $2.369\pm0.08$\\
LSGAN & $2.425 \pm 0.10$\\
WGAN-clipping & $2.771 \pm 0.14$\\
\textbf{MLGAN-clipping} & $2.903\pm 0.14$\\
WGAN-gradient penalty & $3.129 \pm 0.20$\\
\textbf{MLGAN-center penalty} & \textbf{3.296} $\pm$ \textbf{0.17}\\
\hline
\end{tabular*}
\end{table}

As has been stated above, we use {\it Inception Score} to assess the quality of our images and we report the highest score of each model during the training. The result of our experiments is shown in Table \ref{t1}. As we could see from the table, our MLGANs achieve superior performance compared with other models. 

We also show the generated images for each GAN model in Figure \ref{f3}. From the figure we could know that even though for this architecture the attempt to training DCGAN has failed, MLGAN could still generate realistic images, which in part demonstrates the stability of MLGANs.

\subsection{MLGAN on CIFAR-10}
CIFAR-10 dataset \cite{krizhevsky2009learning} consists of $60,000$ $32\times32$ color images in $10$ classes. Again, we use the same model architecture with WGAN with gradient penalty and {\it Inception score} to assess the quality of generated images.

\begin{table}[h]
\caption{\label{t2}Experimental Results on CIFAR-10}
\begin{tabular*}{8cm}{cc}
\hline
Model (same architecture) & Inception Score\\
\hline
DCGAN & $1.030\pm 0.01$\\
\textbf{MLGAN-vanilla} & $3.791\pm 0.45$\\
WGAN-clipping & $4.683 \pm 0.38$\\
\textbf{MLGAN-clipping} & $5.233 \pm 0.29$\\
LSGAN & $5.650\pm0.32$\\
WGAN-gradient penalty & $6.069\pm0.33$\\
\textbf{MLGAN-center penalty} & \textbf{6.279} $\pm$ \textbf{0.33}\\
\hline
\end{tabular*}
\end{table}

As we could see from Figure \ref{f4}, DCGAN has failed in this task again and our model still obtain superior results compared with other GAN models in terms of {\it Inception score}. Also, from Table \ref{t2} we can see that our MLGANs can achieve higher {\it Inception Score}. Furthermore, even though vanilla MLGAN perform the worst, it still do not collapse during the training.

\subsection{Improved Stability}
One of the benefits of MLGANs is that we can train the critic till optimality and the better the critic is; the more reasonable objective function the generator would get. Therefore, the problem of regular GANs, which is we cannot train the discriminator too well, is no longer an issue.

Also, it should be noted that for MLGANs, we just use the same architecture with other GAN models and still get superior results, which could demonstrate the robustness and potential of MLGANs.

Last but not least, even though vanilla MLGAN perform the worst in most cases, it never collapses like regular GAN (DCGAN). In this regard, MLGANs indeed increase the stability of training GANs.

\section{Conclusion}
In this work, we propose a novel framework for generating models named MLGANs, which inherits the adversarial process from GANs but significantly change the goal of both discriminator and generator. To be specific, the discriminator of MLGANs aims at learning an appropriate metric between the real and fake samples and the goal of the generator of MLGANs is to minimize the distance between real and fake samples based on the newly learned metric. 

We also highlight a major issue for training GANs, which is that training GANs is delicate and unstable. In our experiment, we demonstrate that not only do MLGANs achieve superior results compared with other models, but also it is more stable to train MLGANs.

We hope this proposed method could provide readers with a new perspective towards GANs and inspire others to come up with better idea.

\section{Future Work}
Although MLGANs have demonstrated satisfactory results as shown above, there are still several possible improvements that may lead to better performance. First, the objective function for vanilla MLGAN is worth being investigated and by borrowing idea from more state-of-the-art distance metric learning methods, we believe the results would be better. 

Second, to constrain the output of the discriminator of MLGAN, here we propose two possible solutions, namely weight clipping and center penalty. However, these are not the only and the best ways to constrain the discriminator and we encourage researchers to 
find out more possibilities. 

Third, adding label information into MLGANs would be more reasonable since it would be more natural to minimize distance within each class and maximize distance between different classes. 
\bibliographystyle{aaai}
\bibliography{formatting-instructions-latex-2018}
\end{document}